\documentclass[journal]{IEEEtran}

\ifCLASSINFOpdf

\else

\fi

\usepackage{color}
\usepackage{mathptmx}       
\usepackage{helvet}         
\usepackage{courier}        
\usepackage{type1cm}        
\usepackage{makeidx}         
\usepackage{graphicx}        
\usepackage{multicol}        
\usepackage[bottom]{footmisc}
\usepackage{color}
\usepackage{subfigure}
\usepackage{multirow}
\usepackage{booktabs}
\usepackage{caption}

\graphicspath{	{./figures/}}

\begin{document}

\title{Optimize transfer learning for lung diseases in bronchoscopy using a new concept: sequential fine-tuning}

\author{Tao~Tan, Zhang~Li, Haixia~Liu, Ping~Liu, Wenfang~Tang, Hui~Li, Yue~Sun, Yusheng~Yan, Keyu~Li, Tao~Xu, Shanshan~Wan, Ke~Lou, Jun~Xu, Huiming~Ying, Quchang Ouyang, Yuling~Tang, Zheyu~Hu, and Qiang~Li
\thanks{T. Tan is with Department of Biomedical Engineering,
Eindhoven University of Technology, Eindhoven 5600 MB, The Netherlands }
\thanks{Z. Li is with College of Aerospace Science and Engineering,National University of Defense Technology, Changsha 410073, China }
\thanks{H. Liu is with School Of Computer Science, University of Nottingham Malay
sia Campus, Jalan Broga, 43500 Semenyih, Selangor Darul Ehsan, Malaysia}
\thanks{Y. Sun is with Department of Electronic Engineering,
Eindhoven University of Technology, Eindhoven 5600 MB, The Netherlands  }
\thanks{P Liu, W Tang, H Li, Y Yan, K Li, T Xu, S Wan, K Lou, J Xu,and Y Tang are with the First Hospital of Changsha City, Changsha, 410000, China  }
\thanks{H. Ying is with the First Affiliated Hospital of Hunan University of Medicine, Hunan, 418000, China  }
\thanks{Q. Ouyang and Z. Hu is with Department of Breast Medical Oncology, the Affiliated Cancer Hospital of Xiangya Medical School, Central South University, No.283 Tongzipo Road, Changsha 410000, China, email: huzheyu@hnszlyy.com  }
\thanks{Qiang Li is with Department of Respiratory Medicine, Shanghai People's Hospital, Shanghai Tongji University School of Medicine, No.100 Hongqiao Road, Shanghai 20008, China  }
\thanks{Y. Tang, Z. Hu and Q. Li  are corresponding authors}
}

\markboth{Journal of \LaTeX\ Class Files,~Vol.~13, No.~9, September~2014}%
{Shell \MakeLowercase{\textit{Tan et al.}}: Computer-aided diagnosis of lung disease in bronchoscopy}

\maketitle

\begin{abstract}
Bronchoscopy inspection as a follow-up procedure from the radiological imaging  plays a key role in lung disease diagnosis and determining treatment plans for the patients. Doctors needs to make a decision whether to biopsy the patients timely when performing bronchoscopy. However, the doctors also needs to be very selective with biopsies as biopsies may cause uncontrollable bleeding of the lung tissue which is life-threaten. To help doctors to be more selective on biopsies and provide a second opinion on  diagnosis, in this work, we propose a computer-aided diagnosis (CAD) system for lung diseases including cancers and  tuberculosis (TB). The system is developed based on transfer learning. We propose a novel transfer learning method: sentential fine-tuning . Compared to traditional fine-tuning methods, our methods achieves the best performance. We obtained a overall accuracy of 77.0\% a dataset of 81 normal cases, 76 tuberculosis cases and 277 lung cancer cases while the other traditional transfer learning methods achieve an accuracy of 73\% and 68\%. . The detection accuracy of our method for cancers, TB and normal cases are 87\%, 54\% and 91\% respectively. This indicates that the CAD system has potential to improve lung disease diagnosis accuracy in bronchoscopy and  it also might be used to be more selective with biopsies.

\end{abstract}


\IEEEpeerreviewmaketitle

\section{Introduction}

Lung cancer is also called bronchiogenic carcinoma, because about 95\% primary pulmonary cancer is originated from bronchial mucosa. Lung cancer is the top deadly cancer, with the five-year survival rate of 18.1 percent (based on 2017-2013 SEER database). In 2014, there were an estimated 527,228 people living with bronchial lung cancer in the United States (https://seer.cancer.gov/statfacts/html/lungb.html). In China, lung cancer is the most common cancer and the leading cause of cancer death, especially for men in urban areas \cite{Zheng2016}. There were 546,259 tracheal, bronchus, and lung (TBL) cancer deaths, about one third of the 1,639,646 deaths on a global scale in 2013 \cite{Fitzmaurice2015}. Another serious health problem in developing countries derived from lung is tuberculosis (TB). China accounts for more than 10\% of the global TB burden. Currently, Chinese government aims to suppress the TB prevalence from 390 per 100,000 population to 163 per 100, 000 population and stabilize it by 2050 (WHO goal) \cite{Xu2017}. 

Chest x-ray is a cheap and fast imaging device which are commonly used for diagnosis of lung disease including pneumonia, tuberculosis, emphysema and cancer. It is particularly useful for emergency use. With a very small dose of radiation, it generates a 2D projection image including lungs. However, due to its limitation in visualizing lung in 3D, it was gradually replaced by chest CT for lung nodule detection. The downside of the chest CT is its relatively higher radiation. Still in developing country, chest x-ray is used as primary tool for tuberculosis screening or diagnosis. With these radiological imaging tools, radiologists are able to diagnose diseases in clinical or make a referral in a screening situation. 

\begin{figure}[ht]%
\centering
\subfigure[]{%
\includegraphics[height=2in]{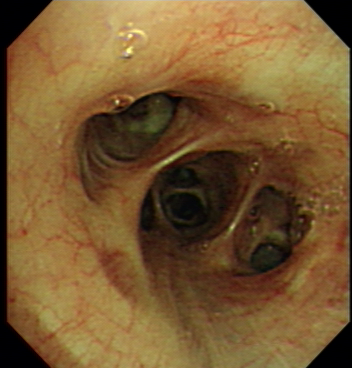}%
}
\subfigure[]{%
\includegraphics[height=2in]{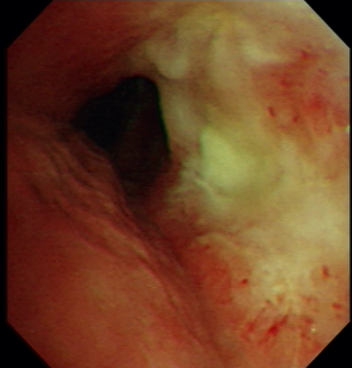}%
}
\subfigure[]{%
\includegraphics[height=2in]{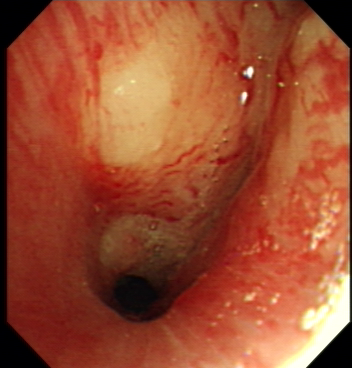}%
}
\caption{A bronchoscopy image of normal (a), TB (b) and cancer (c)}
\label{fig:goodexample}%
\end{figure}

Once patients are suspected to have lung cancer or TB with X-ray or CT, bronchoscopy is followed-up from radiological imaging. Bronchoscopy is used as one of the invasive tool to directly detect the disease since 1960s \cite{Andersen1965}. Fig. \ref{fig:goodexample} shows an example of normal tissue, TB and cancer. In bronchoscopy, usually we can observe that invasive TB, the lumen surface suffers from inflammatory change with hyperemia, edema and ulceration. Lung adenocarcinomas grow extraluminally and lead to lumen stenosis without affecting mucosal surface of lumen. Therefore, the mucosal surface of lumen is relatively smooth. However, squamous lung cancers always form intruding nodules and are difficult to be differentiated from TB granuloma visually. Computational aid is therefore needed in bronchoscopy, especially for lesion discrimination and targeting. Accurate targeting the disease area could significantly reduce the biopsy trauma and increase diagnostic accuracy \cite{Poletti2014}. 

One typical computer-aided diagnosis (CAD) technique for bronchoscopy is so called virtual bronchoscopy (VB) \cite{Vining1996,Summers1997}. VB is normally created from CT scans and used for guiding the bronchoscopy to locate lesions \cite{Reynisson2014,Higgins2015}. Several techniques, such as segmentation \cite{Summers1996,Eberhardt2010}, registration \cite{Mori2002,Wegner2006,Appelbaum2011,Bricault1998} and tracking\cite{Solomon2000,Rai2008,Mori2001}, were introduced to the VB to facilitate the guiding process. Despite the guiding, VB also improves diagnostic accuracy for peripheral lesions compared with traditional bronchoscopy \cite{Reynisson2014}

Due to the limitation of detecting small lesions (e.g. a few cells thick) by traditional  bronchoscopy,  autofluorescence bronchoscopy (AFB) \cite{Gabrecht2007,Zaric2013,Tremblay2016} and narrow band imaging (NBI) \cite{Herth2009} were adopted. These new imaging techniques improved the sensitivity \cite{Sun2011} or specificity \cite{Herth2009} for the early stage cancer detection.  Although AFB and NBI have their advantages for lung cancer diagnosis, the traditional bronchoscopy is still the most used technique in daily clinical routine practice and bronchoscopic biopsy is a cornerstone in the lesion diagnosis. However, bronchoscopic biopsy may cause massive bleeding during the operation and it is life-threating. Therefore, it is very necessary to be selective on the  bronchoscopic biopsies. To improve the diagnosis accuracy without performing bronchoscopic biopsies, CADs technique can play a role.

To the best of our knowledge, CADs was not properly studied for the traditional bronchoscopy. In this study, we are the first to develop a computer-aided diagnosis system to classify cancer, TB and normal tissues in traditional bronchoscopy. We adopted the latest deep learning technique and are the first to propose a novel transfer-learning concept: sequential fine-tuning.

\section{Method}
\section{Convolutional neural networks}
Convolutional neural networks (CNNs) \cite{krizhevsky2012imagenet} is a powerful tool for automatically classifying 2D or 3D image patches (input). It usually contains several pairs of a convolution layer and a pooling layer. The intermediate outputs of these layers are fully connected to a multi-layer perception neural network. Recently, new tricks include  dropout \cite{srivastava2014dropout}, batch normalization \cite{ioffe2015batch} and resnetblock \cite{He2016Identity}. The purpose of dropout is to solve over-fitting caused by co-adaptations during training. Dropout technique improves neural networks performance. Batch normalization helps to accelerate the training of deep networks by normalizing activations. Batch Normalization  achieved  the  same  accuracy  with  14 times fewer training steps\cite{ioffe2015batch}, and outperformed the original model. It also improved the best published result on ImageNet classification using an ensemble  of  batch-normalized  networks. Resnet block was proposed in \cite{He2016Identity}, where they found out that identity shortcut connections and identity after addition activation are important for smoothing information  propagation. They also designed  1000-layer  deep networks and obtained better classification accuracy. Their novel deep networks were designed for creating a direct path  for  propagating information through the entire network instead of within one residual unit and were trained easily in comparison with the original ResNet architechture \cite{He2016Deep}.


\section{Using pretrained networks / transfer learning}
In many medical image classification cases, the number of labeled data are limited for training. Transfer learning has been proposed \cite{pan2010survey,Sinno2010A} to effectively tackle the problem of availability of the labeled data. Transfer learning literally means that experience gained from one subject can be transfered to other subjects. In neural networks, it means that the parameters trained on one dataset can be reused for a new dataset. Usually, transfer learning is used for training a base network and then its first n layers are copied to the first n layers of a new network. The remaining layers of the new network are initialized randomly and trained according to the new task \cite{Yosinski2014How}.



To perform transfer learning, we can keep all layers before the last output layer and connect these layers to a new layer for the new classification problem. To train the networks from the new dataset, We either allow the parameters from the fully connect layers of the networks to be tuned or optimized. The other choice is to fine-tune more layers or even the whole pretrained network layers. It is also possible to keep the first convolutional layer fixed as this layer is often used for edge extraction which is common for all kinds of problems. Since not all parameters are retrained or trained from scratch, the transfer learning is beneficial to problems with a small labeled dataset which is common in medical imaging field. 

Shin et al. \cite{shin2016deep} studied two specific CAD problems, thoraco-abdominal lymph node (LN) detection and interstitial lung disease (ILD) classification. They achieved the state-of-the-art performance on the mediastinal LN detection. They stated that their CNN model analysis can be extended to the design of high performance CAD systems for other medical imaging tasks. Shin et al. \cite{Hoochang2016Deep} managed to solve limited labeled medical data and lacking of domain knowledge problems by utilizing transfer learning techniques. They learned a codebook from 15 million images from ImageNet in  an  unsupervised learning fashion, which encode the fundamental features of those images without experts knowledge. A weighting vector for each image in their experiment was obtained from the codebook and was input into  SVM classifiers for supervised learning. They concluded that using  transfer representation learning to analyze medical data is promising. In comparison, Christodoulidis et al. \cite{Christodoulidis2016Multi} pretrained networks on six public texture datasets and further fune-tuned the network architecture on the lung tissue data. The resulting conversational features are fused with the original knowledge, which was compressed back to the network. Their results showed that the proposed method improved the performance by 2\% compared to the same network without using transfer learning. In \cite{tajbakhsh2016convolutional}, experiments were conducted with the research question "Can the use of pre-trained deep CNNs with sufficient fine-tuning eliminate the need for training a deep CNN from scratch". They concluded that deeply fine-tuned CNNs are useful for analyzing medical images and they performed as well as fully trained CNNs. When the training data is limited, the fine-tuned CNNs even outperformed the fully trained ones.

\section{Our system}
\begin{figure*}%
\centering
\captionsetup{justification=centering}
\includegraphics[width=0.80\textwidth]{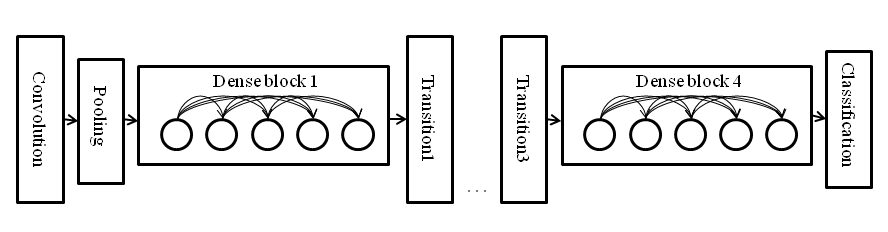}%
\caption{Demonstration of Densenet 121 used in our system.}
\label{fig:densenet121}%
\end{figure*}

In this work, we took a pretrained DenseNet as our pretrained model. Huang et al. \cite{huang2016densely} proposed DenseNet is a network architecture where each layer is directly connected to every other layer in a feed-forward fashion (within each dense block). For each layer, the feature maps of all preceding layers are treated as separate inputs whereas its own feature maps are passed on as inputs to all subsequent layers. In their work, this connectivity pattern yields state-of-the-art accuracies on CIFAR10/100 (with or without data augmentation) and SVHN. On the large scale ILSVRC 2012 (ImageNet) dataset, DenseNet achieves a similar accuracy as ResNet, but using less than half the amount of parameters and roughly half the number of FLOPs. Figure. {~\ref{fig:densenet121}} shows the architecture of this work. The number 121 corresponds to the number of layers with trainable weights (exclude batch norm), for example, convolutional layers and fully connected layers. The additional 5 layers include the initial 7x7 convolutional layer, 3 transitional layers and a fully connected layer.

Inspired by previous work on transfer learning, in this paper we propose a novel way of performing transfer learning. To make best use of limited data meanwhile also to have the power of all layers of the networks, instead of only fine-tuning (FT) the fully connected (FC) layers or fine-tuning the whole networks, we sequentially fine-tuning (SFT) a pretrained network from FC layers to layers before. For a following epoch or a set of epochs, we additionally allow fine-tuning $s$ layers prior to layers which are already fine-tuned in previous epochs. For example, suppose we use $n$ epochs for training and our networks consists of $m$ layers, for each sequential step, we perform $x$ epochs. In total, to train the whole networks, we need to perform $n$/($x$*$m$) steps of SFT. $n$, $x$ and $m$ are parameters to set. 


Since the percentage of each class is different during training, to eliminate the effect of the unbalanced data, we use weighted cross entropy as the cost function to update the parameters of our network. To obtain the final label or class of each sample, we just assign the label or class where the corresponding node in the last layer gives the highest likelihood value. For training, we limit the number of epochs ($n$ ) to be 150 and set $x$ to be 5. To be able to fit our inputs to the required inputs of the pre-trained model, we resize all input images to be 224 by 224.

\section{Materials}
A total of 434 patients who were suspected to have lung diseases by CT/X-ray images were enrolled in this study. All patients were processed at the Department of Respiration in the First Hospital of Changsha City, from January 2016 to November 2017. Inclusion criteria included: 1) image-suspected lung /bronchus diseases; 2) aged between 18 and 70 years old; 3) liver, renal and blood test showed neutrophil count $>$ 2.0 g/l, Hb $>$ 9 g/l, pallete count $>$ 100 g/l, AST and ALT $>$ 0.5 ULN, TBIL $<$1.5 ULN, and Cr $<$ 1.0 ULN. The exclusion criteria included: 1) patients with immune-deficiency or organ-transplantation history; 2) patients with severe heart disease or heart abnormalities, such as cardiac infarction, or severe cardiac arrhythmia. This study was approved by the Ethics Committee of the First Hospital of Changsha City. Informed consent was obtained from each patient before study. Basic demographic and clinical information, including age, sex, image records and treatment history were recorded. 
  
Before bronchoscopy was performed, patients were not allowed to eat and drink for at least 12 hours. 5-10 minutes before operation, patients received 2 percent lidocaine (by high pressure pump) plus localized infiltrating anesthesia. Some received additional conscious sedation or general anesthesia. During the operation, flexible biferoptic bronchoscopy (Olympus BF-260) was inserted to nosal cavaty, glottis and bronchus. Computer workstation is configured to receive bronchoscopy images. Once abnormality of suspect was detected visually, the area was captured by a camera from a high-definition television (HDTV) system and saved as JPG or BMP files (319 by 302 pixels). 
  
Biopsy was the gold standard for diagnosing malignant/premalignant airway disease. Therefore, specimens from bronchial biopsy were obtained in all cases of this study. Specimens for cytologic / pathologic diagnosis were obtained from the following ways: brushing from the lesion or bronchial washings, fine needle aspiration biopsy, and forcep biopsy from visible tumor or TB lesions. Histological diagnosis was made by experienced pathologists. Two independent pathologists firstly made their diagnosis individually. If their diagnosis were inconsistent, another arbitrator pathologist would make the decision. Such histological results are used as ground truth of this study. According to pathological confirmation, among recruited 434 patients, 81 cases were diagnosed as healthy, 76 were diagnosed as TB, and 277 were diagnosed as lung cancer patients. In total, we collected 81 normal cases, 277 lung cancer cases and 76 TB cases.

\section{Experiments and Results}
\subsection{Experiments}

In this work, considering the limited number of samples, in order to obtain an unbiased evaluation of the classification performance, a 2-fold cross-validation is employed to evaluate the performance of our method. Specifically, the input dataset is randomly divided into 2 equal parts, where one part is left for testing, and the other part is split again for training (70\%) and validation (30\%), to avoid the bias. The best classifier based on the validation set is used for testing. Such a procedure is repeated 2 times with a different part used for testing. We pooled the results from both two parts and evaluate the performance measurements. 

Since we aim to solve a three-class classification problem, the measurements of accuracy (ACC) and confusion matrix are used for the evaluation purpose. We also tackle with the the two-class (binary) classification problems such as abnormal versus normal cases, TB versus cancer cases and non-cancer cases versus cancer cases. The receiver operating characteristic (ROC) analysis and the area under the ROC curve (AUC) are used for the evaluating the two-class classification performance. 

\subsection{Results}


\begin{figure}[ht]%
\centering
\subfigure[Confusion matrix]{%
\label{fig:confusionBest}%
\includegraphics[height=2.8in]{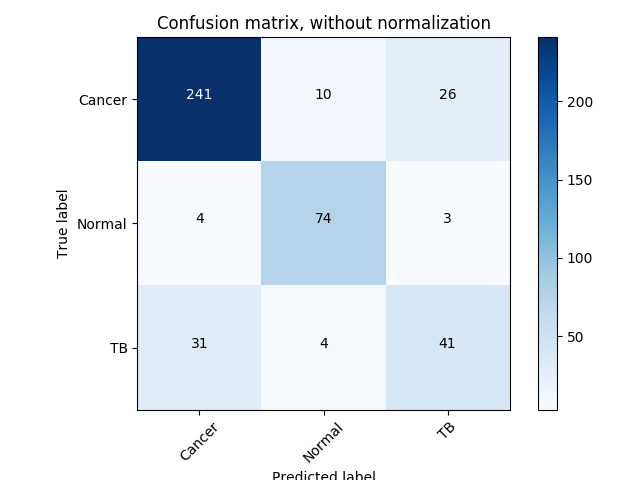}}%
\qquad
\subfigure[Normalized confusion matrix]{%
\label{fig:confusionNormBest}%
\includegraphics[height=2.8in]{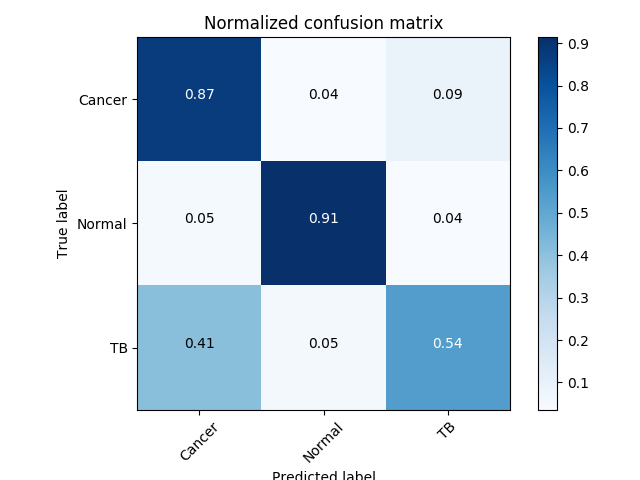}}%
\caption{The confusion matrix and the normalized confusion matrix of our prosed method using sequential fine-tuning}
\label{fig:confusion3}%
\end{figure}

Fig. {~\ref{fig:confusionBest}} and Fig. {~\ref{fig:confusionNormBest}} show the confusion matrix and the normalized confusion matrix from fine-tuning all layers together, fine-tuning the fully connected layers and our proposed method (sequential fine-tuning) respectively. It can be observed that in general, our proposed method gives the most accurate result. The overall accuracy of the three methods are 73.7\%, 70.2\% and 82.0\%, respectively.

\begin{figure}[ht]%
\centering
\includegraphics[height=2.8in]{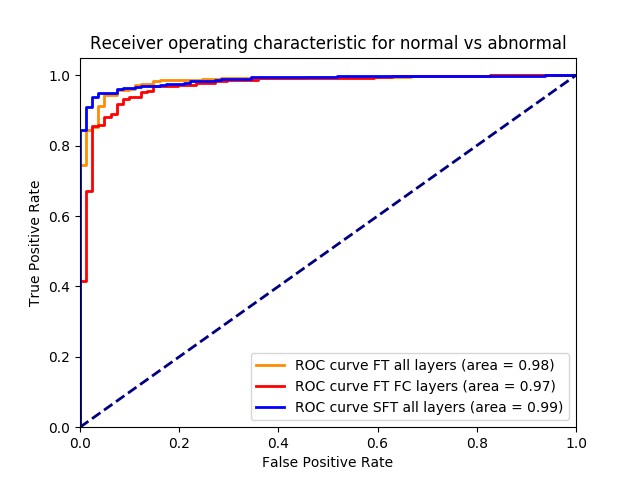}%
\caption{ROC curve of binary situations of abnormal (TB+cancer, class 1) and normal (class0) with fine-tuning ally layers together (orange),  only fine-tuning the fully connected layers (red), our prosed method using sequential fine-tuning (blue), respectively  }
\label{fig:roc11}%
\end{figure}

Fig. {~\ref{fig:roc11}} shows the receiver operating characteristic curve (ROC) for the binary situation between normal cases and abnormal cases (TB+cancer) from fine-tuning all layer together, fine-tuning the fully connected layers and our proposed method (sequential fine-tuning) respectively. The area under the ROC curve is 0.98, 0.97 and 0.99, respectively.

\begin{figure}[ht]%
\centering

\centering
\includegraphics[height=2.8in]{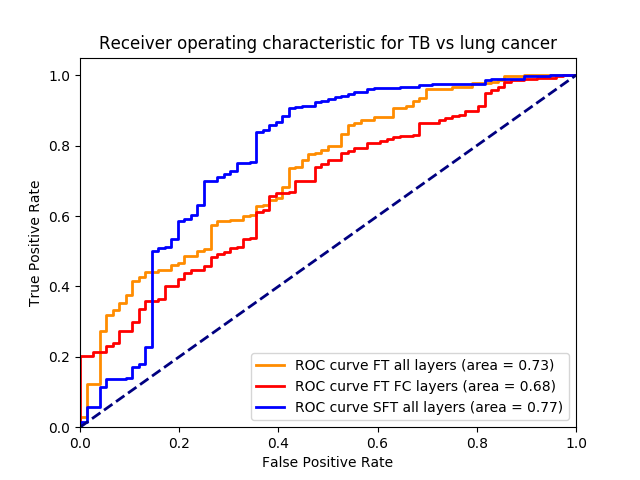}%
\caption{ROC curve of binary situations of TB (class 0) and cancer (class 1) with fine-tuning ally layers together (orange),  only fine-tuning the fully connected layers (red), our prosed method using sequential fine-tuning (blue), respectively.}
\label{fig:roc31}%
\end{figure}

Fig. {~\ref{fig:roc31}} shows the receiver operating characteristic curve (ROC) for the binary situation between TB cases and cancer cases from fine-tuning all layers together, fine-tuning the fully connected layers and our proposed method (sequential fine-tuning) respectively. The area under the ROC curve is 0.73, 0.68 and 0.77, respectively.

\begin{figure}[ht]%
\centering

\centering
\includegraphics[height=2.8in]{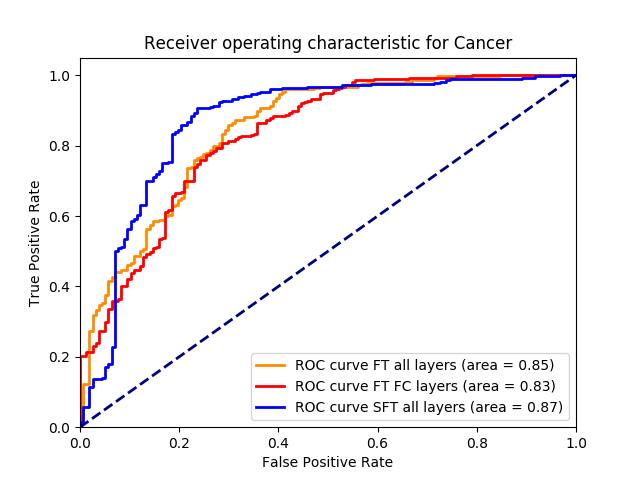}%
\caption{ROC curve of binary situations of non-cancer (class 0) and cancer (class 1) with fine-tuning ally layers together (orange),  only fine-tuning the fully connected layers (red), our prosed method using sequential fine-tuning (blue), respectively.}
\label{fig:roc41}%
\end{figure}

Fig. {~\ref{fig:roc41}} shows the receiver operating characteristic curve (ROC) for the binary situation between non-cancer cases and cancer cases from fine-tuning all layers together, fine-tuning the fully connected layers and our proposed method (sequential fine-tuning) respectively. The area under the ROC curve is 0.85, 0.83 and 0.87, respectively.

\begin{figure}[ht]%
\centering
\subfigure[]{%
\includegraphics[height=2in]{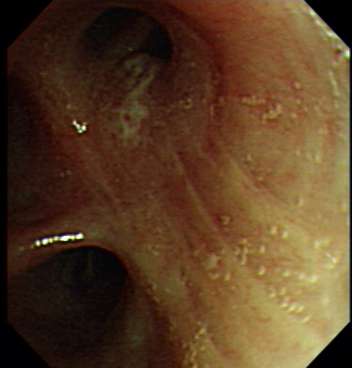}%
}
\subfigure[]{%
\includegraphics[height=2in]{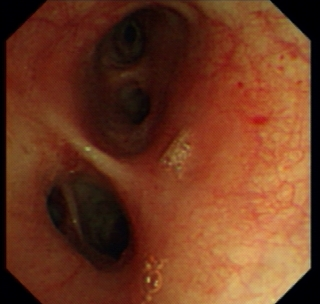}%
}
\subfigure[]{%
\includegraphics[height=2in]{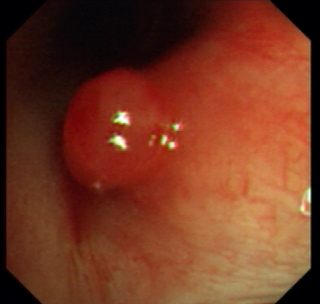}%
}
\caption{A cancer case was misclassified as a normal case (a), a normal case was misclassified as a cancer case (b), a TB case was misclassified as a cancer case(c)}
\label{fig:example}%
\end{figure}

Figure. {~\ref{fig:example}} shows examples of misclassified cases. Figure.{~\ref{fig:example}}a indicates a cancer case that has pale mucosa and yellow secretion and was mis-classified as a normal case by CAD, while Figure.{~\ref{fig:example}}b indicates a normal case that has smooth red mucosa and was mis-classified as a cancer case by CAD. Figure.{~\ref{fig:example}}c. indicates a TB case that has a round nodule with smooth mucosa and was mis-classified as a cancer case by CAD. Mucosa color, secretions and smoothness are important features for cancer discrimination. In cancer cases, the tumor mucosa is pale, rigid and has dirty secretion. The mis-classification in Figure.{~\ref{fig:example}} might be due to a small dataset for training. Larger training set would extract more minor mucosa features to avoid such mis-classification.

\begin{table}[h]
\begin{tabular}{| l | l | l | l | l |}
  \hline			
  method & ACC & AUC prob. 1 & AUC prob. 2 & AUC prob. 3  \\  
  FT all layers  & 0.73 & 0.98  & 0.73 & 0.85\\
  FT FC layers  & 0.68  & 0.97 & 0.68 & 0.83\\
  SFT all layers  & \textbf{0.77} & \textbf{0.99} & \textbf{0.77} & \textbf{0.87}\\
  \hline 

\end{tabular}
\caption[]{Performance measures including three-class classification accuracy, AUC for problem 1 of abnormal versus normal cases, problem 2 of AUC for cancer versus TB cases and problem 3 of non-cancer versus cancer cases from different methods }
\label{table:measures}
\end{table}

Table \ref{table:measures} summarizes different performance measures from different methods. Our proposed methods outperforms other compared methods regarding all measures.

\section{Conclusion and discussion}
A computer aided diagnosis system was developed for the
classification of normal, tuberculosis and lung cancer cases in bronchoscopy. In the system, a deep learning model based on pre-trained DenseNet is applied. Using the sequential fine tuning, our model in combination with 2-fold-cross-validation, obtained a overall accuracy of 82.0\% a dataset of 81 normal cases, 76 tuberculosis cases and 277 lung cancer cases. The detection accuracy for cancers, TB and normal cases were 87\%, 54\% and 91\% respectively. This indicates that the CAD system has potential to improve diagnosis and that it also might be used to be more selective with biopsies. Furthermore, we showed that the performance of the deep-learning model was improved with our proposed sequential fine-tuning.

To our best knowledge, we are the first to bring up the concept of sequential fine-tuning in deep learning networks and we showed the benefits of using sequential fine-tuning compared to fine-tuning all layers and fine-tuning only fully connected layers. Our explanation is that since the dataset size is small, it is not reasonable to fine-tune a very large set of parameters of the whole networks at the beginning. Therefore, we choose to sequentially and gradually fine-tune more and more layers from a pre-trained model. The other benefits of doing sequential fine-tuning is that instead of fitting data to two sub-models of the DenseNet (a model with no layers fixed and a model with fully connected layers fixed), we fit our data to more sub-models as sequentially we fixed different sets of layers. By doing so, we have a better chance of finding a good model for the data.

We also investigated our classification power of different binary classification situations. The area under the ROC curve from the binary classification of abnormal cases and normal cases is very high (0.99). From the ROC curve, we can see that we can keep the sensitivity of detecting abnormal cases of our CAD system to be 1 while the specificity is 0.65. It means that our CAD system can identify 65\% normal cases without missing any abnormal cases. It has the potential to reduce the false positive rate of doctors and avoid further with biopsies of these normal patients. The area under the ROC curve from the binary classification of TB cases and cancer cases is 0.87 where there is still space to improve. Although the discrimination power is not very high, we can still triage these abnormal patients and almost 10\% of TB patients are correctly identified by our CAD system without missing any cancer patients. Again, these patients would not necessarily go for biopsies. For some cases, the CAD system did not perform well. Figure. {~\ref{fig:example}} shows misclassified cases. The TB case was mistakenly classified as a lung cancer case by our CAD system. This TB nodule looks very like a malignant tumor. However, for doctors, there is still one feature for discrimination: TB surface was more smooth than cancer surface. With a larger training set, more features would be extracted automatically, and this kind of mistakes would be eliminated or suppressed. In this study, TB cases from the training set is small and thus the trained model is not good enough to differentiate minor features of difficult cases. 

In this study, we investigate neither the actual diagnostic performance of doctors on bronchoscopy images nor the performance of doctors with the aid of our CAD system. In the future, we will conduct a reader study to evaluate the benefits of using our CAD system.
Bronchoscopy as an invasive instrument plays a key role in lung disease diagnosis and determining treatment plans for the patients. With bronchoscopy doctors can directly observe the lung tissue and diagnose the problem to some extent. The doctors needs to make a decision whether to biopsy the patients timely when performing bronchoscopy. However, the doctors also needs to be very selective with biopsies as biopsies can easily cause uncontrollable bleeding of the lung tissue which is life-threaten. With aid of our computer system, doctors can already correctly eliminate 65\%normal patients, 10\% of TB patients to avoid unnecessary biopsies/risk for patients which is of great help in clinical operation. To further suppress the number of biopsies, in the future, we will investigate the possibility of boosting this CAD system for identifying specific types of lung cancers. That means more labeled data should be collected in the future. The future work may also extend our CAD system in combining with other imaging techniques(e.g. AFB) to cover broader a range of diseases and meanwhile combining deep learning networks together with human crafted features from domain knowledge.




\bibliographystyle{unsrt}
\bibliography{ref}

\end{document}